\documentclass[conference]{sty/IEEEtran}

\usepackage{amsmath,amsfonts,amssymb}
\usepackage{algorithmic}
\usepackage{algorithm}
\usepackage{array}
\usepackage[caption=false,font=normalsize,labelfont=sf,textfont=sf]{subfig}
\usepackage{textcomp}
\usepackage{stfloats}
\usepackage{url}
\usepackage{verbatim}
\usepackage{booktabs}
\usepackage{multirow}
\usepackage{rotating}
\usepackage{makecell}
\usepackage{cite}
\usepackage{xspace}
\usepackage{bm}
\usepackage{sty/my_acronyms}
\usepackage{siunitx}
\usepackage{tikz}
\usepackage{sty/my_acronyms}
\usetikzlibrary{shapes,arrows,calc}
\newcommand{\RomanNumeralCaps}[1] {\MakeUppercase{\romannumeral #1}}

\IEEEoverridecommandlockouts

\title{\LARGE \bf
Toward 6-DOF Autonomous Underwater Vehicle Energy-Aware Position Control based on Deep Reinforcement Learning: Preliminary Results
\thanks{(*) These authors contributed equally. This work was supported by the David and Lucile Packard Foundation and the ANID-Chile grants for doctoral program N°21200358 and master program N°22231977.}
}

\author{\IEEEauthorblockN{Gustavo Boré${}^*$}
\IEEEauthorblockA{\textit{Pontificia Universidad Católica de Chile} \\
Santiago, Chile \\
Email: gibore@uc.cl}
\\ 
\IEEEauthorblockN{Sebastián Rodríguez-Martínez}
\IEEEauthorblockA{\textit{Monterey Bay Aquarium Research Institute} \\
Moss Landing, California 95039--9644\\
Email: srodriguez@mbari.org}
\and
\IEEEauthorblockN{Vicente Sufán${}^*$}
\IEEEauthorblockA{\textit{Pontificia Universidad Católica de Chile} \\
Santiago, Chile \\
Email: vicente.sufan@uc.cl}
\\ 
\IEEEauthorblockN{Giancarlo Troni}
\IEEEauthorblockA{\textit{Monterey Bay Aquarium Research Institute} \\
Moss Landing, California 95039--9644 \\
Email: gtroni@mbari.org}
}

\begin{document}

\maketitle
\thispagestyle{empty}
\pagestyle{empty}

\begin{abstract}
The use of \acp{auv} for surveying, mapping, and inspecting unexplored underwater areas plays a crucial role, where maneuverability and power efficiency are key factors for extending the use of these platforms, making \ac{6dof} holonomic platforms essential tools. Although \ac{pid} and \acl{mpc} controllers are widely used in these applications, they often require accurate system knowledge, struggle with repeatability when facing payload or configuration changes, and can be time-consuming to fine-tune. While more advanced methods based on \acl{drl} have been proposed, they are typically limited to operating in fewer degrees of freedom.
This paper proposes a novel \acs{drl}-based approach for controlling holonomic \ac{6dof} \acp{auv} using the \ac{tqc} algorithm, which does not require manual tuning and directly feeds commands to the thrusters without prior knowledge of their configuration. Furthermore, it incorporates power consumption directly into the reward function. Simulation results show that the \acl{tqc-hp} method achieves better performance to a fine-tuned \ac{pid} controller when reaching a goal point, while the \acl{tqc-ea} method demonstrates slightly lower performance but consumes \SI{30}{\percent} less power on average.
\end{abstract}

\acresetall


\section{Introduction}
\label{sec:intro}

\noindent The exploration of our oceans has traditionally relied on expensive, large, and energy-consuming underwater vehicles for tasks such as surveying or mapping unexplored areas. These vehicles have the ability to execute different maneuvers depending on the thruster configuration and design. However, various factors have limited these applications, primarily a lack of precise underwater navigation and maneuverability. Specifically, the maneuverability of these platforms is constrained by either inadequate actuation, which restricts movement to fewer than \ac{6dof}, or by the limitations of the controllers, which involve a complex trade-off between tuning efforts and extensibility to more \acp{dof}. Numerous research vehicles are limited from their nominal \ac{6dof}, being designed to be rolling or pitching stable via buoyancy design and mechanically simpler with fewer thrusters, but at the cost of limiting maneuverability. Furthermore, in some cases, the thruster configuration limits the ability of the vehicle to move independently in all six axes, forcing the system to become nonholonomic.

Even though the control of the \ac{auv} is critical for trajectory tracking and the capabilities of the vehicle to operate in complex terrains, the energy used for this task is also a critical consideration, as \acp{auv} operates under battery power and optimal use leads to better autonomy. For research purposes, deploying a holonomic and power-efficient \ac{auv} can improve efficiency in data collection, seabed mapping, and maneuverability, providing significant advantages over nonholonomic vehicles.

\begin{figure}[t!]
\centering
\includegraphics[width=\linewidth]{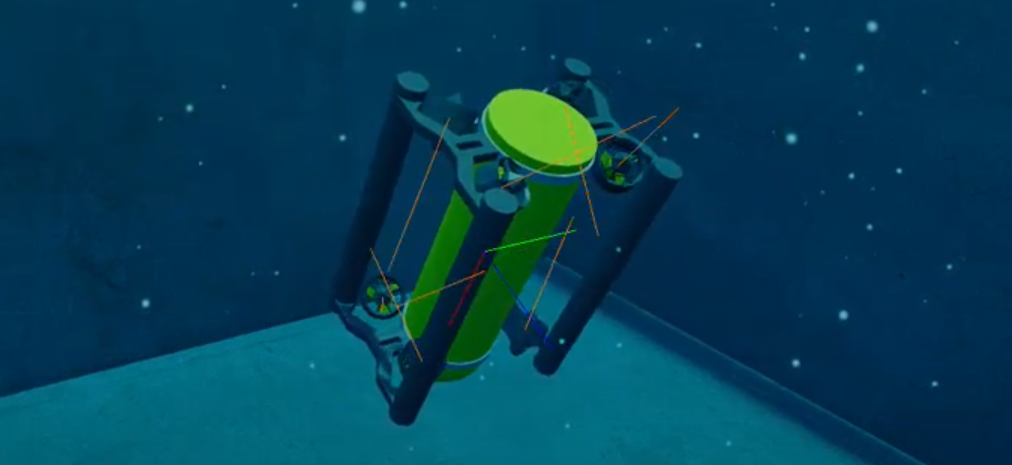}%
\caption{Stonefish model of \acs{mola} \acs{6dof} \acs{auv}, \acs{mbari}'s autonomous research platform for complex terrain exploration.}
\label{fig:mola}
\end{figure}


\acp{auv} exhibit complex behavior in underwater environments and can be mathematically modeled as coupled \ac{6dof} dynamic systems. As such, they are highly nonlinear, and their characterization requires the identification of more than 200 parameters \cite{mcfarland2021} to represent the hydrodynamics of the vehicle, which can be solved either numerically or experimentally. A widely used approach to control an \ac{auv} is the classical \ac{pid} control because of its linear nature; however, it is unable to fulfill optimality for energy-efficient controllers, and its performance is highly affected by the nonlinearities, its tight link to the configuration of the thrusters, the payload, and the speed range. Any change in the above requires adjusting the gains, which can involve up to 36 terms for the widely used velocity and position integrated controller. Even though \acp{nn} have been used to tune the \ac{pid} gains in \cite{hernandez2016}, they were not integrated experimentally.

Similarly, model-based approaches, such as \ac{mpc}, are widely used because they incorporate system dynamics to frame control as an optimization problem, enabling control actions to minimize a cost function and achieve more optimal outcomes. Anderlini et al. \cite{anderlini2019} propose an adaptive \ac{mpc} framework as a complement to a \ac{pid} controller when the \ac{auv} carries a payload, correcting the difference in dynamics caused by an unmodeled mass, while Fernandez and Hollinger \cite{fernandez2016} present a similar approach for controlling an \ac{auv} in ocean waves. Although \ac{mpc} can achieve high performance, it also relies on system identification, which can be highly complex or unavailable, as it requires accounting for all inertial and drag coefficients. Errors in estimating these parameters or linearizing the system for simplification and convexity in cost optimization can lead to a model mismatch with the real system. Consequently, the optimal solution is calculated iteratively, increasing exponentially with additional \acp{dof}, leading to noticeable time delays. To overcome these problems, Martinsen et al. \cite{Martinsen2020} propose a \ac{nn}-based system identification algorithm to create a data-driven \ac{mpc}, solving numerically a structured known equation of motion to generate data and train a \ac{nn} that can successfully approximate the system's dynamics. However, it has been only tested in simple linear simulated systems.

The advances in \ac{drl} have established it as a reliable methodology for analyzing complex systems and building data-driven control algorithms, based purely on the relation of \ac{mimo} systems and the ability to handle large amounts of data, benefiting from the lack of a rigid model structure. The use of \ac{drl} extends to path planning, attitude control, and navigation for \acp{auv} \cite{Singh2023}, focusing on the high-level control of the vehicle. Carlucho et al. \cite{carlucho2018} propose an end-to-end Actor-Critic approach to control a 5-\acs{dof} \ac{auv} in both a simulation environment and a test tank for experimental validation. Their work focuses on speed control for linear velocities without addressing rotational movements. Furthermore, the study lacks a direct comparison of the proposed controller against a \ac{pid} controller regarding position error and power requirements across all scenarios. Notably, the evaluation does not include testing with a \ac{6dof} trajectory that incorporates rotations. Lagattu et al. \cite{lagattu2024} developed a \ac{drl}-based thruster recovery controller, merging a classical \ac{pid} controller and a \ac{drl} backup controller when an undiagnosed fault is detected, based on the \ac{auv} behavior using a \ac{sac} algorithm. Unlike a \ac{pid} controller, the proposed controller can successfully reach a way-point with some overshooting after a failure. Lidtke et al. \cite{lidtke2024} based their work on the surveying task over a vertical cylinder, including dynamic forces computed with \ac{cfd} in a \ac{2d} representation and computing the movement as the solution to a 3-\acs{dof} differential equation. Several algorithms were tested, where \ac{tqc} outperformed other algorithms such as \ac{sac}, \ac{ppo}, \ac{ddpg}, and \ac{td3}. Even though \cite{carlucho2018, lagattu2024, lidtke2024} successfully enable the vehicle to track trajectories, they use the whole power range of the thrusters to achieve a task-efficient controller. Therefore, energy consumption is not addressed as a constraint.

Only a few reported works have covered end-to-end controllers and their energy-efficiency analysis. Huang et al. \cite{Huang2022A} designed a specific reward function with an energy consumption term that penalizes high-value outputs of the control actions, leading to an energy-efficient \ac{sac} policy for an analytically-simulated torpedo-shaped underactuated \ac{auv} in 5-\acs{dof} within a \ac{3d} action space; however, the capability of this algorithm to operate in \ac{6dof} \acp{auv} is not stated. Sola et al. \cite{Sola2022} propose the use of \ac{sac} for the control and guidance of a \ac{6dof} \ac{auv}, which outperforms the \ac{pid} in terms of energy efficiency; however, without including an energy efficiency term explicitly in the reward function. Nevertheless, \cite{Huang2022A} and \cite{Sola2022} achieve power efficiency through trajectory planning instead of the control itself; thus, the energy metrics are a consequence of proper mission planning rather than the controller itself.


To address the open challenges and limitations of previously proposed methods, this paper introduces \ac{tqc-hp} and \ac{tqc-ea}: two end-to-end \ac{drl}-based approaches for the low-level control of a holonomic \ac{6dof} \ac{auv} using the \ac{tqc} algorithm. These methods require neither manual tuning nor prior knowledge of the thruster configuration. To the best of our knowledge, this is the first report of a \ac{drl} algorithm successfully controlling a holonomic \ac{6dof} \ac{auv} in a simulation environment, directly mapping the eight thrusters without manual configuration (\ac{tqc-hp} and \ac{tqc-ea}). Furthermore, it is the first to incorporate energy-awareness to optimize energy consumption within the controller (\ac{tqc-ea}).

The paper is structured as follows. Section \RomanNumeralCaps{2} overviews required concepts. Section \RomanNumeralCaps{3} details the proposed method. Section \RomanNumeralCaps{4} describes the evaluation methodology, while Section \RomanNumeralCaps{5} reports the evaluation and analysis of the results. The results and conclusions are summarized in Section \RomanNumeralCaps{6}.
\section{Background}
\label{sec:background}

\subsection{Deep Reinforcement learning}

\ac{drl} \cite{RichardSutton20} is a method that aims to train an agent's policy $\pi$ to map states into actions by interacting with the environment. This is achieved by maximizing a numerical reward signal and using a \ac{mdp} framework to regulate the interaction between the \ac{rl} agent’s policy and the environment. At each time step, the agent observes a state $\bm{s}$, takes an action $\bm{a}$, and upon transitioning to the next state, receives a reward $r$. Once the episode (i.e., process) is complete, the accumulated reward is calculated as the sum of all time steps rewards in that episode.

\ac{drl} methods can be model-based or model-free. Model-based methods use a model to predict the next state and reward, while model-free methods learn solely from experiencing the unmodeled and unknown consequences of an action. While learning from trial and error may result in less efficient learning, model-free methods have the advantage when a model is unavailable or inaccurate.

\begin{figure}[t!]
\centering
\includegraphics[width=0.45\textwidth]{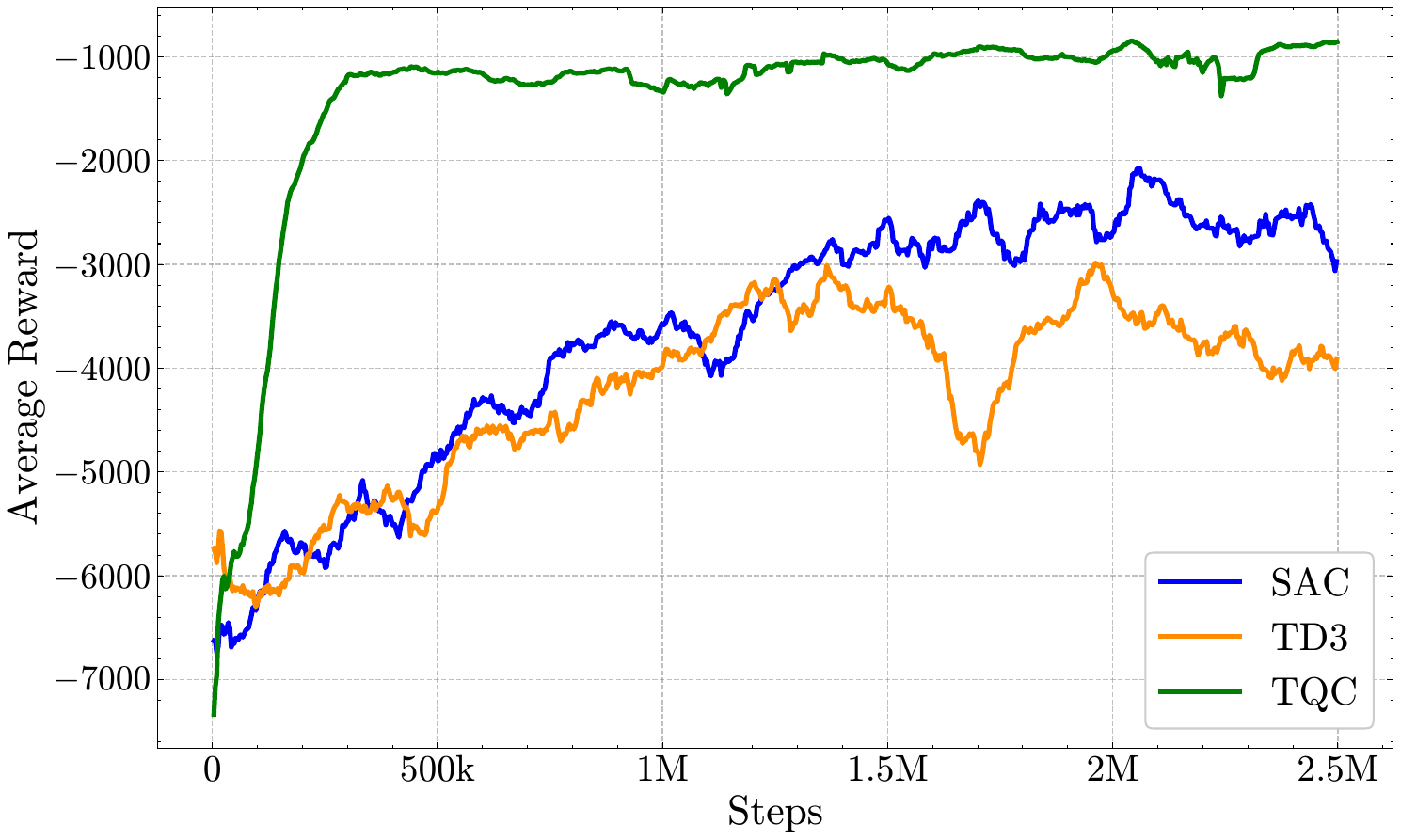}%
\caption{Average reward per episode over a moving window of 100 episodes obtained by the TQC, SAC, and TD3 algorithms during a $2.5\times10^6$ step training, equivalent to 3125 episodes.}
\label{fig:rewards}
\end{figure}

\subsection{\ac{6dof} Error Computation}

The position errors are determined by the difference between the current position $(x, y, z)$ and the goal position $(x_d, y_d, z_d)$ following the North-East-Down (NED) convention, computed as
\begin{equation}
    e_x(t) = x^t - x_d^t,\; e_y(t) = y^t - y_d^t,\; e_z(t) = z^t - z_d^t.
\label{eq:errors}
\end{equation}

To compute the error in attitude, we will evaluate the difference between the current orientation and the goal attitude, both with respect to the fixed world frame. This involves representing both poses as rotation matrices ($\bm{R}\in SO(3)$) and converting their difference to exponential coordinates $[\bm{{e_\theta}}]\in so(3)$ through the matrix logarithm:
\begin{equation}
     [\bm{{e_\theta}}(t)] = \log(\bm{R}(t)^T \cdot \bm{R}_d)
\end{equation}

Then, the skew-symmetric matrix $[\bm{{e_\theta}}(t)]$ is converted into its vector representation $\bm{{e_\theta}}(t) \in \mathbb{R}^3$, where its entries correspond to the element-wise error for the attitude, defined as
\begin{equation}
    \begin{bmatrix} \theta_{x}^t & \theta_{y}^t & \theta_{z}^t \end{bmatrix} = \bm{{e_\theta}}(t).
    \label{eq:attitude_error}
\end{equation}

Furthermore, to provide a single metric for attitude error evaluation, we compute $\theta^t$ based on the axis-angle representation for $\bm{{e_\theta}}(t)$, as described in \eqref{eq:theta_error}. By using this metric, we obtain a global evaluation of orientation, which aligns the controller's performance with practical manual navigation comparisons.
\begin{equation}
    \theta^t = ||\bm{{e_\theta}}(t)||
    \label{eq:theta_error}
\end{equation}

\section{Proposed Approach}
\label{sec:proposed_approach}

In this work, we introduce \acf{tqc-hp} and \acf{tqc-ea}: two end-to-end \ac{drl}-based approaches for the low-level control of a holonomic \ac{6dof} \ac{auv} using the \ac{tqc} algorithm. These methods require neither manual tuning nor prior knowledge of the thruster configuration.

\subsection{Action space}

The action space $\bm{a}(t) \in \mathbb{R}^{8}$, normalized between $[-1, 1]$, is designed to enable the agent to generate precise control commands for the vehicle’s thrusters, ensuring the desired movement and orientation based on the given state inputs. The action space is defined as:
\begin{equation}
\bm{a}(t) = \begin{bmatrix}
T_{1}^t & T_{2}^t & T_{3}^t & T_{4}^t & T_{5}^t & T_{6}^t & T_{7}^t & T_{8}^t
\end{bmatrix}^T,
\label{eq:action_space}
\end{equation}

\noindent where $\bm{a}(t)$ represents the normalized \ac{pwm} signals sent to the eight thrusters at time $t$.

\subsection{Observation space}

The observation space $\bm{s}(t) \in \mathbb{R}^{20}$, normalized between $[-1, 1]$, is designed to capture all relevant environmental information necessary for the agent’s decision-making, enabling an understanding of the system’s current state relative to the desired goal. The observation space is defined as:
\begin{equation}
\bm{s}(t) = \begin{bmatrix}
\bm{e}(t) & \bm{v}(t) & \bm{a}(t-1)
\end{bmatrix}^T,
\label{eq:observation_space}
\end{equation}

\noindent where $\bm{e}(t)=[e_{x}^t, e_{y}^t, e_{z}^t, \theta_{x}^t, \theta_{y}^t, \theta_{z}^t]^T$ represents the error vector at time $t$ between the current pose and the goal pose for each \ac{dof}, $ \bm{v}(t)=[v_{x}^t, v_{y}^t, v_{z}^t, \omega_{x}^t, \omega_{y}^t, \omega_{z}^t]^T$ corresponds to the spatial twist of the vehicle at time $t$, composed by the linear and angular velocities, and $\boldsymbol{a}(t-1)$ the action in the previous time step.

\begin{table}[b!]
\centering
\caption{Weights $\alpha_i$ for reward functions for \ac{tqc-hp} and \ac{tqc-ea}}
\label{tab:weights}
\begin{tabular}{ccccccccc}
\toprule
 & $\bm{\alpha_1}$ & $\bm{\alpha_2}$ & $\bm{\alpha_3}$ & $\bm{\alpha_4}$ & $\bm{\alpha_5}$ & $\bm{\alpha_6}$  \\ \midrule
\textbf{\ac{tqc-hp}} & $-4$ & $-4$ & $-3$ & $-1.8$ & $-1$ & $0$ \\
\textbf{\ac{tqc-ea}} & $-4$ & $-4$ & $-3$ & $-1.7$ & $-0.8$ & $-0.3$  \\ \bottomrule
\end{tabular}
\end{table}

\subsection{Reward function}

To evaluate policy performance and provide feedback during the training stage, an optimal reward function must be defined to enable the model to learn the desired behavior. In this work, we define the reward function \eqref{eq:reward_fx} that aims to bring the \ac{auv} closer to the target position in each of the \ac{6dof}, linearly penalizing the agent for positional \eqref{eq:reward_position} and angular \eqref{eq:reward_attitude} errors over time. Additionally, the reward function aims to generate smoother commands for the thrusters by penalizing signal fluctuations \eqref{eq:reward_smoothness} and to minimize the utilization of each thruster, thereby reducing the energy consumption of the \ac{auv} \eqref{eq:reward_power}. Each term is weighted with the parameter  $\alpha_i \leq 0$ to denote the importance of each term within \eqref{eq:reward_fx}.
\begin{subequations}
\begin{align}
r(t) &= \sum\nolimits_{i=1}^{4} r_i(t) \label{eq:reward_fx} \\
r_1(t) &= \alpha_1 \cdot{|e_x^t|} + \alpha_2 \cdot{|e_y^t|} + \alpha_3 \cdot{|e_z^t|}\label{eq:reward_position} \\
r_2(t) &= \alpha_4 \cdot{|\theta^t|}
\label{eq:reward_attitude} \\
r_3(t) &= \sum\nolimits_{i=1}^{8} \alpha_5 \cdot \left|{T_{i}^t - T_{i}^{t-1}} \right| \label{eq:reward_smoothness} \\
r_4(t) &= \sum\nolimits_{i=1}^{8} \alpha_6 \cdot \left|{T_i^t} \right| \label{eq:reward_power}
\end{align}
\end{subequations}

\subsection{Proposed \ac{drl} algorithms}

The proposed approaches are based on a \ac{drl} framework using a \ac{tqc} algorithm \cite{kuznetsov2020}. The selection of this algorithm was guided by an experiment conducted to compare the performance of the \ac{tqc}, \ac{sac}, and \ac{td3} algorithms in a \ac{6dof} \ac{auv} position control problem. The methodology described in Section \ref{sec:metodology} was used, and the three methods were trained using the same reward function to compare which of the aforementioned models obtained the highest average reward during the training episodes, indicating superior performance in the specific control task.

Fig. \ref{fig:rewards} shows the average reward of an episode over a moving window of 100 episodes obtained by the three \ac{drl} algorithms during training. It is evident that the \ac{tqc} algorithm converges in a smaller number of episodes compared to \ac{sac} and \ac{td3}, as well as reaching higher reward values than the other two algorithms. This experiment correlates with the results obtained by Lidtke et al. \cite{lidtke2024} for a 3-\acs{dof} \ac{auv} and confirms our decision to use the \ac{tqc} algorithm for the task of controlling an \ac{auv} in \ac{6dof}.

In this work, two different approaches of \ac{tqc} models are proposed: \ac{tqc-hp} aims to reach the goal position in the \ac{6dof} using smooth commands to the thrusters, i.e., its reward function includes the components $r_1$, $r_2$, and $r_3$, while \ac{tqc-ea} has the same objectives as \ac{tqc-hp} but adds energy-awareness, minimizing the energy consumption of the \ac{auv} by incorporating the $r_4$ component in its reward function. Table \ref{tab:weights} presents the $\alpha_i$ weights used in the reward function for both approaches. Both approaches are implemented in \textit{Python3} using the \textit{Stable Baselines3} library, which is based on principles used in both \ac{sac} and \ac{td3} algorithms but utilizes the distributed representation of a critic, truncation of critic predictions, and a set of multiple critics.

\section{Evaluation Methodology}
\label{sec:metodology}

The proposed methods have been developed in the numerical simulation environment Stonefish  \cite{cieslak2019}, which allows the introduction of underwater physics and environmental conditions together with the vehicle geometry, accounting for the complete vehicle dynamics, including drag effects and added masses. The middleware used to communicate between the vehicle's sensors and the control algorithm is established through the \ac{lcm} library \cite{LCM}, which is the current middleware used by the field platform that can be used for future field testing; therefore, easing the transition from simulation to reality.

\begin{figure}[t!]
\centering
\includegraphics[width=0.48\textwidth]{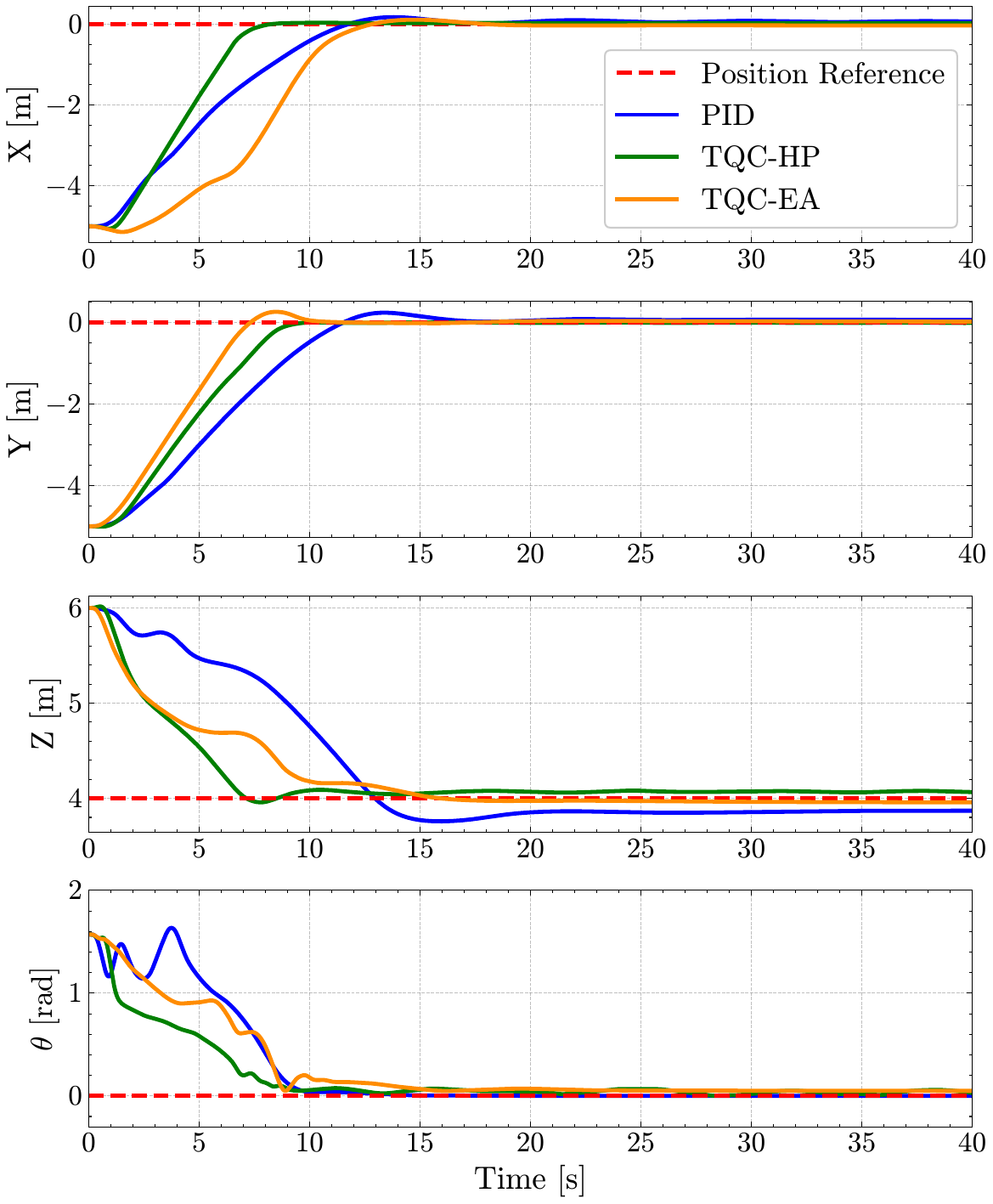}%
\caption{Position in the $x$, $y$, and $z$ axes and angular distance to the target, $\theta$, over time during one of the evaluation episodes.}
\label{fig:position_vs_time}
\end{figure}

The field platform modeled within the simulation environment (depicted in Fig. \ref{fig:mola}) is \ac{mola} \ac{6dof} \ac{auv}, a research platform used for developing and testing novel research methods to enable the exploration of deep and rugged terrains. This platform is owned and operated by the \acs{compas} Lab at the \acf{mbari}. This system is equipped with an \ac{imu}, \ac{dvl}, depth sensor, cameras, sonar, and eight thrusters, and can perform complete holonomic movements in \ac{6dof}. Currently, \ac{mola} has a fine-tuned double-loop 36-gain \ac{pid} controller, with good performance in the same simulation environment. It has been extensively validated in the physical platform during tank tests and is used as the proper performance comparison for the \ac{drl} controller.

Both \ac{tqc-hp} and \ac{tqc-ea} are trained within the simulation environment in over $2.5\times10^6$ steps, starting from a random pose, without restriction in orientation but constrained to the region defined by an interior box with sides of $\SI{3}{\meter}$ and an exterior box of sides $\SI{6}{\meter}$, both centered on the goal point. The goal point is located at $x = \SI{0}{\meter}$, $y = \SI{0}{\meter}$, $z = \SI{4}{\meter}$, $r = \SI{0}{\radian}$, $p = \SI{0}{\radian}$, and $h = \SI{0}{\radian}$. The \ac{auv} is required to reach the goal point within $\SI{40}{\second}$, i.e., 800 steps, with a sampling time of $\SI{50}{\milli\second}$. 

In particular, for the energy consumption evaluation, addressing the relation between thruster power and action commands performed by the \ac{drl} controller is relevant. Based on the Blue Robotics T200 thrusters datasheet \cite{T200Thruster}, we can associate the power draw for a given \ac{pwm}, being \SI{1500}{} equivalent to $0 \; RPM$ and $\left[\SI{1100}{}, \SI{1900}{}\right]$ the bounds for forward and backward thrust, respectively, which can be mapped to the normalized actions space; therefore we can infer the energy consumption directly from the PWM signal, which can be approximated to a 4th order polynomial.

\begin{figure*}[t!]
\centering
\includegraphics[width=0.95\textwidth]{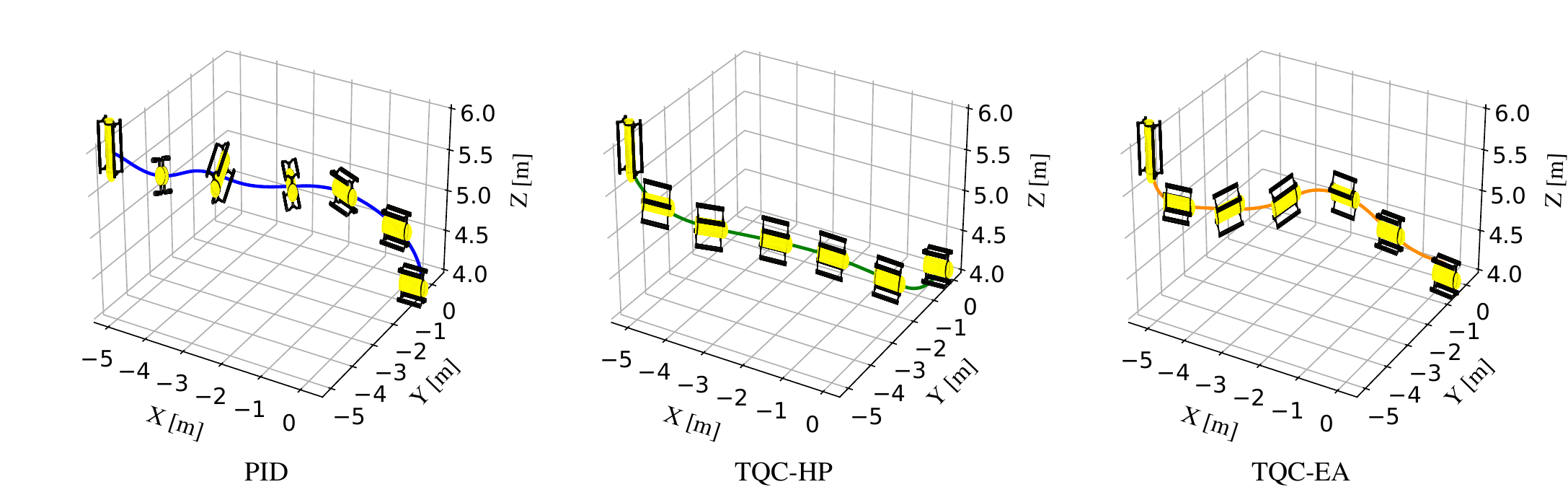}%
\caption{3D trajectory followed by the \ac{auv} in the same episode as depicted in Fig. 3, using the PID, \ac{tqc} HP, and \ac{tqc} EA controllers}
\label{fig:3d_trayectory}
\end{figure*}

For the final evaluation of the trained policy, both proposed methods using the \ac{pid} controller as a benchmark were tested, combining three starting points for each \ac{dof}, resulting in $3^6 = 729$ combinations, all targeting $x = \SI{0}{\meter}$, $y = \SI{0}{\meter}$, $z = \SI{4}{\meter}$, $r = \SI{0}{\radian}$, $p = \SI{0}{\radian}$, and $h = \SI{0}{\radian}$. The starting positions for each axis are: $x = \SI[parse-numbers=false]{\{-5, 0, 5\}}{\meter}$, $y = \SI[parse-numbers=false]{\{-5, 0, 5\}}{\meter}$, and $z = \SI[parse-numbers=false]{\{2, 4, 6\}}{\meter}$; the initial attitudes are: $r = \SI[parse-numbers=false]{\{0, \pi/2, -\pi/2\}}{\radian}$, $p = \SI[parse-numbers=false]{\{0, \pi/2, -\pi/2\}}{\radian}$, and $h = \SI[parse-numbers=false]{\{0, \pi/2, -\pi/2\}}{\radian}$.

Both training and testing were conducted on an Intel i7 12th gen processor, 16 GB of RAM, and an NVIDIA RTX 3050 4 GB graphics card. To assess the performance of the methods, we calculated the average power consumption of the episodes, the \ac{rmse} error in position and orientation, and the approximated settling time ($t_s$).

\section{Results and Discussion}
\label{sec:results}

The trained agents \ac{tqc-hp} and \ac{tqc-ea} are tested against the tuned \ac{pid} controller in 729 episodes, using the \ac{rmse} to assess performance in pose and energy consumption. Results are summarized in Table \ref{tab:rmse}.

\begin{table}[b!]
\centering
\caption{RMSE and standard deviation results for $x$, $y$, $z$, $\theta$, and $t_s$ in 729 episodes. The best result in each column is bolded}
\label{tab:rmse}
\begin{tabular}{l@{\hspace{3.5mm}}c@{\hspace{3.5mm}}c@{\hspace{3.5mm}}c@{\hspace{3.5mm}}}
\toprule
 & \textbf{\ac{pid}} & \textbf{\ac{tqc-hp}} & \textbf{\ac{tqc-ea}} \\
\midrule
$\bm{x}$\textbf{-axis} (\SI{}{\meter}) & $0.99 \pm 0.73$ & $\mathbf{0.96 \pm 0.64}$ & $1.14 \pm 0.74$ \\
$\bm{y}$\textbf{-axis} (\SI{}{\meter}) & $1.01 \pm 0.74$ & $\mathbf{1.00 \pm 0.66}$ & $1.18 \pm 0.76$ \\
$\bm{z}$\textbf{-axis} (\SI{}{\meter}) & $0.50 \pm 0.32$ & $\mathbf{0.36 \pm 0.19}$ & $0.42 \pm 0.26$ \\
$\bm{\theta}$ (\SI{}{\radian}) & $0.36 \pm 0.15$ & $\mathbf{0.32 \pm 0.08}$ & $0.41 \pm 0.14$ \\
$\bm{t_s}$ (\SI{}{\second}) & $15.67 \pm 9.52$ &  $\mathbf{8.28 \pm 1.95}$ & $21.95 \pm 11.15$ \\
\bottomrule
\end{tabular}
\end{table}

The \ac{tqc-hp} controller exhibited superior performance across all evaluated metrics. In comparison to the \ac{pid} controller, \ac{tqc-hp} achieved a reduction in position \ac{rmse} by \SI{3.03}{\percent} (\SI{0.03}{\meter}) on the x-axis, \SI{0.1}{\percent} (\SI{0.01}{\meter}) on the y-axis  and \SI{28}{\percent} (\SI{0.14}{\meter}) on the z-axis. Additionally, in terms of attitude \ac{rmse}, \ac{tqc-hp} demonstrated a notable improvement of \SI{11.1}{\percent} (\SI{0.04}{\radian}).

Conversely, the \ac{tqc-ea} controller, which incorporates energy consumption constraints, showed slightly lower performance relative to both the \ac{pid} and \ac{tqc-hp} controllers. Specifically, when compared to the \ac{pid} controller, \ac{tqc-ea} increased the position \ac{rmse} by \SI{15.2}{\percent} (\SI{0.15}{\meter}) on the x-axis and \SI{16.8}{\percent} (\SI{0.17}{\meter}) on the y-axis, although it achieved a \SI{16.0}{\percent} (\SI{0.08}{\meter}) improvement on the z-axis. The attitude \ac{rmse} also deteriorated by \SI{13.89}{\percent} (\SI{0.05}{\radian}).

Finally, comparing the two proposed methods, \ac{tqc-hp} and \ac{tqc-ea}, reveals further insights. The \ac{tqc-ea} controller underperformed against \ac{tqc-hp}, with position \ac{rmse} increasing by \SI{13.27}{\percent} (\SI{0.13}{\meter}) in the x-axis, \SI{9.90}{\percent} (\SI{0.1}{\meter}) in the y-axis, and \SI{29.4}{\percent} (\SI{0.1}{\meter}) in the z-axis. In terms of attitude \ac{rmse}, the decline was \SI{21.85}{\percent} (\SI{0.08}{\radian}).

From Fig. \ref{fig:position_vs_time}, we can analyze the \ac{auv}'s pose over time for one of the 729 evaluated episodes. In this episode, the \ac{auv} starts at a position of $x = \SI{-5}{\meter}$, $y = \SI{-5}{\meter}$, $z = \SI{6}{\meter}$, $r = \SI[parse-numbers=false]{-\pi/2}{\radian}$, $p = \SI[parse-numbers=false]{\pi/2}{\radian}$, and $h = \SI[parse-numbers=false]{-\pi/2}{\radian}$, which corresponds to an angle $\theta = \SI[parse-numbers=false]{\pi/2}{\radian}$. The goal position is $x = \SI{0}{\meter}$, $y = \SI{0}{\meter}$, $z = \SI{4}{\meter}$, $r = \SI{0}{\radian}$, $p = \SI{0}{\radian}$, and $h = \SI{0}{\radian}$, with a corresponding angle $\theta = \SI{0}{rad}$.

We observe that the \ac{tqc-hp} controller reaches the references in less time (\SI{8.28}{\second}) compared to the \ac{pid} (\SI{15.67}{\second}) and \ac{tqc-ea} (\SI{21.95}{\second}) controllers. Additionally, the \ac{tqc-ea} controller exhibits a smoother response over time due to the energy and smoothness constraints applied. Despite these constraints, it manages to reach the goal position in a time comparable to that of the \ac{pid} controller.

\begin{figure}[t!]
\centering
\includegraphics[width=0.41\textwidth]{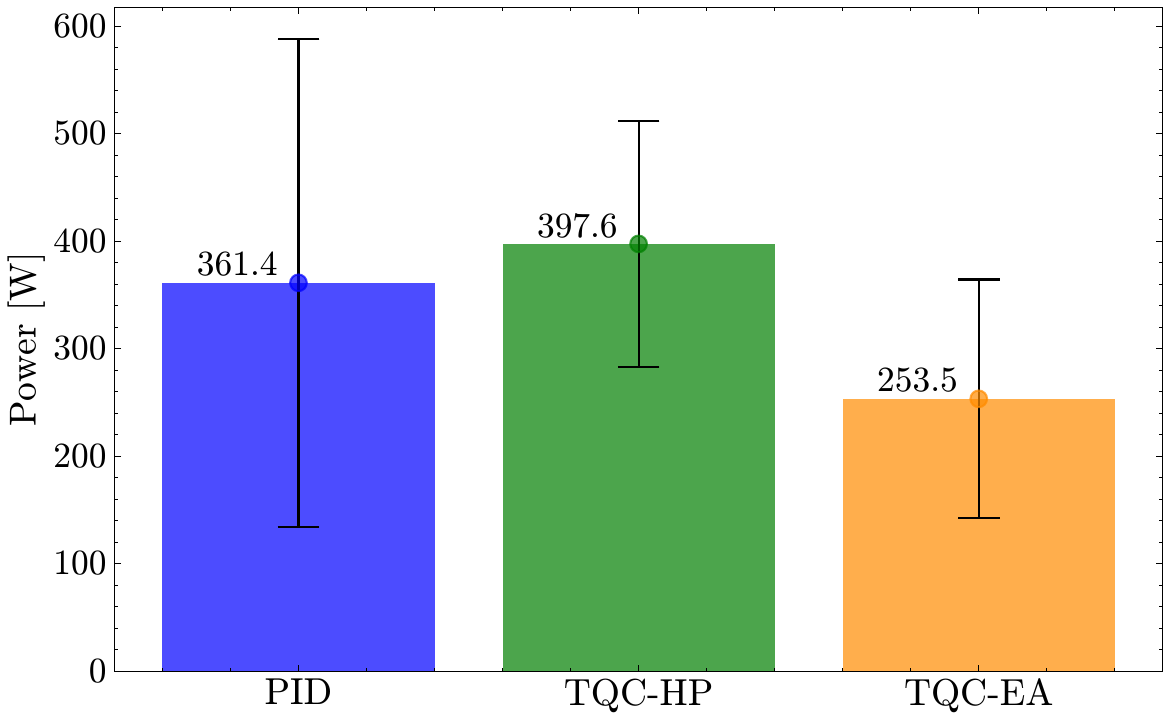}%
\caption{Average power consumption and standard deviation of the \ac{auv} during the evaluation using \ac{pid}, \ac{tqc-hp} and \ac{tqc-ea} controllers.}
\label{fig:power}
\end{figure}

Regarding power consumption, Fig. \ref{fig:power} shows the average power used by the \ac{auv} during the evaluation episodes. The \ac{tqc-hp} controller, which has no restrictions on thruster usage, consumed \SI{10}{\percent} more power than the \ac{pid} controller. In contrast, the \ac{tqc-ea} controller, which incorporates restrictions on thruster usage, used \SI{30}{\percent} less power than the \ac{pid} controller during the evaluation episodes.

We can also visualize the \ac{3d} trajectory followed by the \ac{auv} in the same episode for the three controllers in Fig. \ref{fig:3d_trayectory}. The \ac{pid} controller begins its trajectory by rotating while moving towards the goal position, deviating from the angular reference during the initial part of the trajectory before ultimately reaching the goal. The \ac{tqc-hp} controller demonstrates smoother angular alignment in the initial meters, consistently approaching the angular reference; however, the path taken is not the most intuitively direct. The \ac{tqc-ea} controller's trajectory is similar to that of the \ac{pid} controller but is achieved with \SI{30}{\percent} less power.

Furthermore, the tuning of the weight for the reward function in the \ac{tqc-ea} controller can be further optimized depending on the application's priorities. By adjusting these weights, the power consumption can be reduced further, potentially at the cost of increased pose error, and vice versa.

The results indicate that while the \ac{tqc-hp} controller provides the best positional accuracy and alignment with the goal angular position, the \ac{tqc-ea} model offers a balance between accuracy and energy efficiency. This balance makes the \ac{tqc-ea} model particularly suitable for applications where power consumption is critical, such as long-range cruising, which is typical for \acp{auv}. Furthermore, the energy savings are achieved through the controller itself rather than through path planning, as seen in previous methods.

\section{Conclusions}
\label{sec:conclusions}

The use of \acp{auv} for surveying, mapping, and inspecting unexplored underwater areas plays a crucial role, where maneuverability and power efficiency are key factors for extending the use of these platforms. Advances in \ac{drl} have established it as a reliable methodology for analyzing complex systems and developing data-driven control algorithms.

In this paper, we introduce \ac{tqc-hp} and \ac{tqc-ea}, two end-to-end \ac{drl}-based approaches for the low-level control of a holonomic \ac{6dof} \ac{auv} using the \ac{tqc} algorithm. These methods require no manual tuning or prior knowledge of the thruster configuration and incorporate energy awareness (via \ac{tqc-ea}). Using a simulation environment and the model of \ac{mola}, we trained and demonstrated that the proposed methods show promise compared to the widely used \ac{pid} controller. In particular, the energy-aware method, utilizing a suitable reward function, demonstrates well-balanced performance between behavior and power consumption, which can be further fine-tuned. In future work, we plan to evaluate and deploy the algorithm with the physical platform in the field at \ac{mbari} and perform specific tasks for science operations.

\section*{Acknowledgments}

The authors gratefully acknowledge the \acs{compas} laboratory at \acs{mbari} engineering and technical team.


\addtolength{\textheight}{-8.75cm}   

\bibliography{refs/IEEEabrv, refs/DRL}
\bibliographystyle{IEEEtran}

\end{document}